\documentclass[11pt,a4paper]{article}
\usepackage[hyperref]{emnlp-ijcnlp-2019}
\usepackage{times}
\usepackage{latexsym}

\usepackage{graphicx}
\usepackage{url}
\usepackage{amsmath}
\usepackage{amssymb}
\usepackage{empheq}
\usepackage{framed}
\usepackage{array}
\usepackage{bm}
\usepackage{microtype}
\usepackage{booktabs}

\newcommand{\PreserveBackslash}[1]{\let\temp=\\#1\let\\=\temp}
\newcolumntype{C}[1]{>{\PreserveBackslash\centering}p{#1}}
\newcolumntype{R}[1]{>{\PreserveBackslash\raggedleft}p{#1}}
\newcolumntype{L}[1]{>{\PreserveBackslash\raggedright}p{#1}}

\newif\ifcomments
\commentstrue 
\ifcomments
    \newcommand{\BB}[1]{{\color{red}(BB: #1)}}
    \newcommand{\jb}[1]{{\color{blue}(JB: #1)}}
    \newcommand{\matt}[1]{{\color{olive}(MG: #1)}}
\else
    \providecommand{\BB}[1]{}
    \providecommand{\jb}[1]{}
    \providecommand{\matt}[1]{}
\fi

\newcommand{\sV}{\mathcal{V}}

\newcommand{\sU}{\mathcal{U}}

\newcommand{\softmax}{\text{softmax}}
\newcommand{\spider}{\textsc{Spider}}

\aclfinalcopy 

\title{Global Reasoning over Database Structures for Text-to-SQL Parsing}

\author{Ben Bogin$^{1}$ ~~~~~ Matt Gardner$^{2}$ ~~~~~
Jonathan Berant$^{1,2}$ \\
\mbox{}\\
$^1$School of Computer Science, Tel-Aviv University \\
$^2$Allen Institute for Artificial Intelligence \\
\small{\texttt{ben.bogin@cs.tau.ac.il, mattg@allenai.org, joberant@cs.tau.ac.il}}}

\date{}

\begin{document}
\maketitle

\begin{abstract}
State-of-the-art semantic parsers rely on auto-regressive decoding, emitting one symbol at a time.
When tested against complex databases that are unobserved at training time (zero-shot), the parser often struggles to select the correct set of database constants in the new database, due to the local nature of decoding.
In this work, we propose a semantic parser that globally reasons about the structure of the output query to make a more contextually-informed 
selection of database constants. We use message-passing through a graph neural network to softly select a subset of database constants for the output query, conditioned on the question. Moreover, we train a model to rank queries based on the global alignment of database constants to question words.
We apply our techniques to the current state-of-the-art model for \spider{}, a zero-shot semantic parsing dataset with complex databases, increasing accuracy from 39.4\% to 47.4\%.
\end{abstract}
\section{Introduction}
The goal of zero-shot semantic parsing \cite{krishnamurthy2017neural,xu2017sqlnet,yu2018spider,yu2018syntaxsqlnet,herzig2018zeroshot} is to map language utterances into executable programs in a new environment, or database (DB). The key difficulty in this setup is that the parser must map new lexical items to DB constants that weren't observed at training time.

Existing semantic parsers 
handle this mostly through a \emph{local} similarity function between words and DB constants, which considers each word and DB constant in isolation.
This function is combined with an auto-regressive decoder, where
the decoder chooses the DB constant that is most similar to the words it is currently attending to. Thus, selecting DB constants is done one at a time rather than as a set, and  informative global considerations are ignored.

\begin{figure}
    \centering
    \includegraphics[scale=0.27]{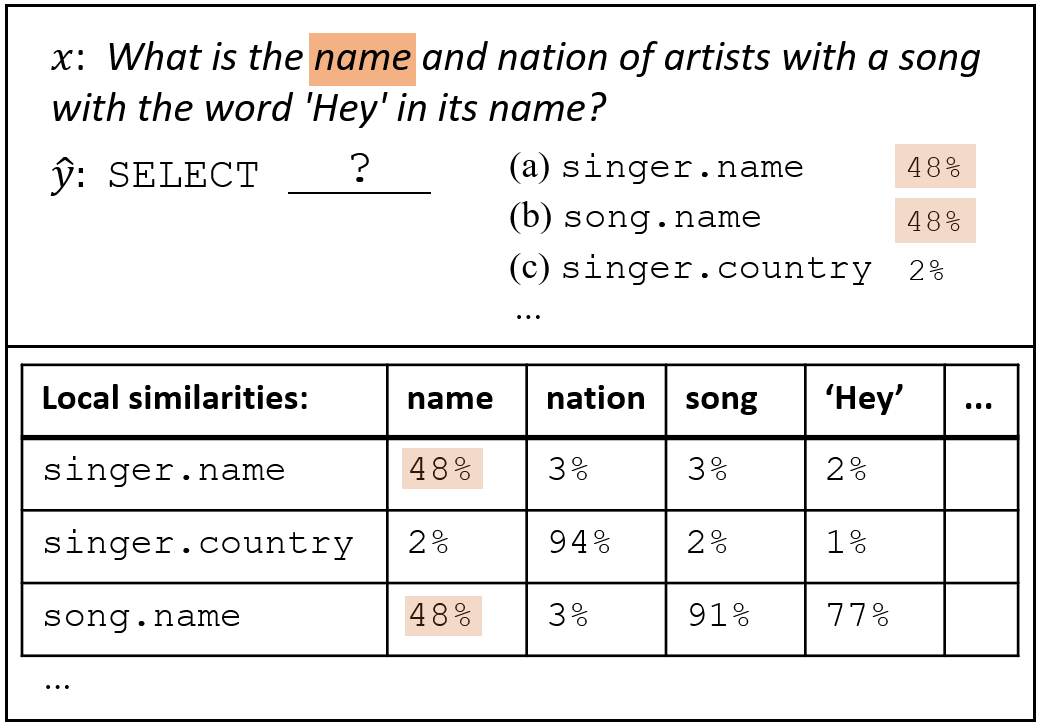}
    \caption{An example where choosing a DB constant based on local similarities is difficult, but the ambiguity can be resolved through global reasoning (see text). }

    \label{fig:problem}
\end{figure}

Consider the example in Figure~\ref{fig:problem}, where a question is mapped to a SQL query over a complex DB. After decoding \texttt{SELECT}, the decoder must now choose a DB constant. Assuming its attention is focused on the word \emph{`name'} (highlighted), and given local similarities only, the choice between the lexically-related DB constants (\texttt{singer.name} and \texttt{song.name}) is ambiguous. However, if we globally reason over the DB constants and question, we can combine additional cues. 
First, a subsequent word \emph{`nation'} is similar to the DB column \texttt{country} which belongs to the table \texttt{singer}, thus selecting the column \texttt{singer.name} from the same table is more likely. Second, the next appearance of the word \emph{`name'} is next to the phrase \emph{'Hey'}, which appears as the value in one of the cells of the column \texttt{song.name}. Assuming a one-to-one mapping between words and DB constants, again \texttt{singer.name} is preferred.

\label{sec:model}

In this paper, we propose a semantic parser that reasons over the DB structure and question to make a \emph{global} decision about which DB constants should be used in a query. We extend the parser of \newcite{bogin2019gnn}, which learns a representation for the DB schema at parsing time. First, we perform message-passing through a graph neural network representation of the DB schema,
to softly select the set of DB constants that are likely to appear in the output query.
Second, we train a model that takes the top-$K$ queries output by the auto-regressive model and re-ranks them based on a global match between the DB and the question.
Both of these technical contributions can be applied to any zero-shot semantic parser. 

We test our parser on \spider{}, a  zero-shot semantic parsing dataset with complex DBs. We show that both our contributions improve performance, leading to an accuracy of 47.4\%, well beyond the current state-of-the-art of 39.4\%.

Our code is available at \url{https://github.com/benbogin/spider-schema-gnn-global}.
\section{Schema-augmented Semantic Parser}
\label{sec:base_model}

\noindent
\textbf{Problem Setup}
We are given a training set $\{(x^{(k)}, y^{(k)}, S^{(k)})\}_{k=1}^N$, where $x^{(k)}$ is a question, $y^{(k)}$ is its translation to a SQL query, and $S^{(k)}$ is the schema of the corresponding DB. We train a model to map question-schema pairs $(x,S)$ to the correct SQL query. Importantly, the schema $S$ was not seen at training time.

A DB schema $S$ includes : (a) A set of DB tables, (b) a set of columns for each table, and (c) a set of foreign key-primary key column pairs where each pair is a relation from a foreign-key in one table to a primary-key in another. Schema tables and columns are termed \emph{DB constants}, denoted by $\sV$.

We now describe a recent semantic parser from \newcite{bogin2019gnn}, focusing on the components relevant for selecting DB constants.

\noindent
\textbf{Base Model}
The base parser is a standard top-down semantic parser with grammar-based decoding \cite{xiao2016sequence,yin2017syntactic,krishnamurthy2017neural,rabinovich2017abstract,lin2019grammar}. The input question $(x_1, \dots, x_{|x|})$ is encoded with a BiLSTM, where the hidden states $\bm{e}_i$ of the BiLSTM are used as contextualized representations for the word $x_i$. The output query $y$ is decoded top-down with another LSTM using a SQL grammar, where at each time step a grammar rule is decoded. Our main focus is decoding of DB constants, and we will elaborate on this part.

The parser decodes a DB constant whenever the previous step decoded the non-terminals \texttt{Table} or \texttt{Column}. To select the DB constant, it first computes an attention distribution over the question words $\{\alpha_i\}_{i=1}^{|x|}$ in the standard manner \cite{bahdanau2015neural}. Then the score for a DB constant $v$ is $s_v = \sum_{i} \alpha_i s_\text{link}(v, x_i)$, where $s_\text{link}$ is a local similarity score, computed from learned embeddings of the word and DB constant, and a few manually-crafted features, such as the edit distance between the two inputs and the fraction of string overlap between them. The output distribution of the decoder is simply $\softmax(\{s_v\}_{v \in \sV})$. Importantly, the dependence between decoding decisions for DB constants is weak -- the similarity function is independent for each constant and question word, and decisions are far apart in the decoding sequence, especially in a top-down parser.

\begin{figure*}
    \centering
    \includegraphics[scale=0.42]{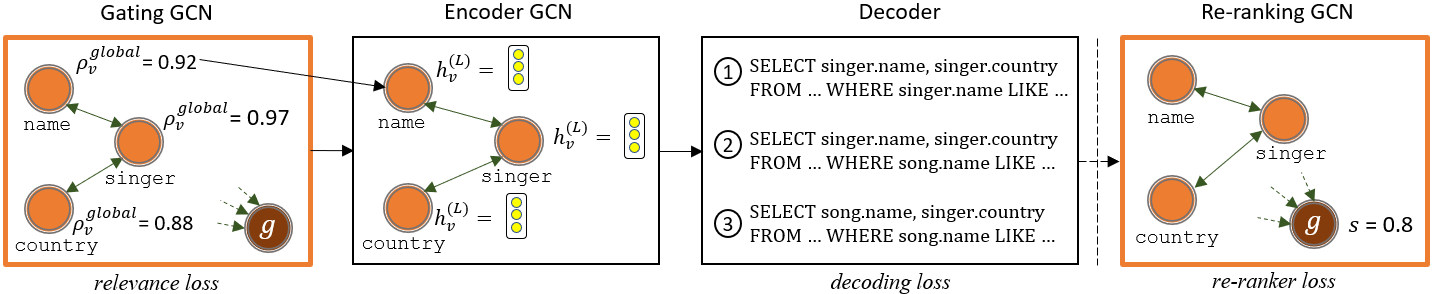}
    \caption{High-level overview, where our contributions are in thick orange boxes. First, a relevance score is predicted for each of the DB constants using the gating GCN. Then, a learned representation is computed for each DB constant using the encoder GCN, which is then used by the decoder to predict $K$ candidates queries. Finally, the re-ranking GCN scores each one of these candidates, basing its score only on the selected DB constants. The dashed line and arrow indicate no gradients are propagated from the re-ranking GCN to the decoder, as the decoder outputs SQL queries. Names of loss terms are written below models that are trained with a loss on their output.
    }
    \label{fig:highlevel}
\end{figure*}

\noindent
\textbf{DB schema encoding}
In the zero-shot setting, the schema structure of a new DB can affect the output query. To capture DB structure, \newcite{bogin2019gnn} learned a representation $\bm{h}_v$ for every DB constant, which the parser later used at decoding time. This was done by converting the DB schema into a graph, where nodes are DB constants, and edges connect tables and their columns, as well as primary and foreign keys (Figure~\ref{fig:highlevel}, left). A graph convolutional network (GCN) then learned representations $\bm{h_v}$ for nodes end-to-end \cite{Cao2018QuestionAB,sorokin-gurevych-2018-modeling}.

To focus the GCN's capacity on important nodes, a \emph{relevance probability} $\rho_v$ was computed for every node, and used to ``gate" the input to the GCN, conditioned on the question. Specifically, given a learned embedding $\bm{r}_v$ for every database constant, the GCN input is $\bm{h}^{(0)}_v  =  \rho_v \cdot \bm{r}_v$. Then, the GCN recurrence is applied for $L$ steps. At each step, nodes re-compute their representation based on the representation of their neighbors, where different edge types are associated with different learned parameters \cite{li2016gated}.
The final representation of each DB constant is $\bm{h}_v = \bm{h}_v^{(L)}$.

Importantly, the relevance probability $\rho_v$, which can be viewed as a soft selection for whether the DB constant should appear in the output,
was computed based on local information only: First, a distribution $p_\text{link}(v \mid x_i) \propto \exp(s_\text{link}(v, x_i))$  was defined, and then $\rho_v = \max_i{p_\text{link}(v \mid x_i)}$ was computed deterministically. Thus, $\rho_v$ doesn't consider the full question or DB structure. We address this next.

\section{Global Reasoning over DB Structures}
\label{sec:model}

\begin{figure}
    \centering
    \includegraphics[scale=0.32]{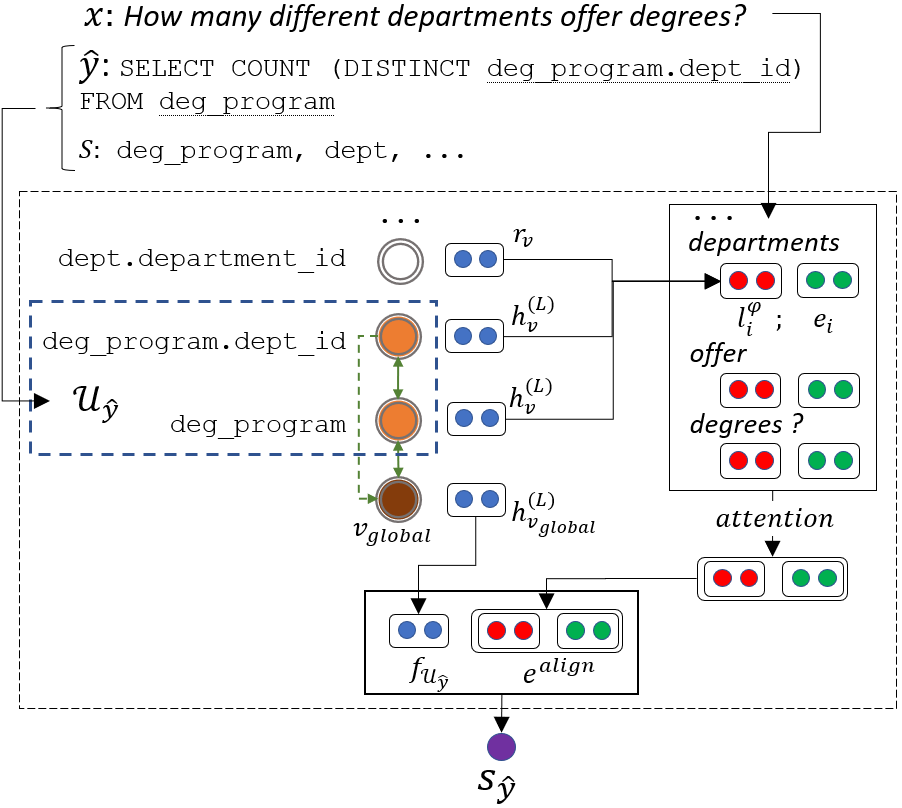}
    \caption{The re-ranking GCN architecture (see text).}
    \label{fig:reranker}
\end{figure}

Figure~\ref{fig:highlevel} gives a high-level view of our model, where the contributions of this paper are marked by thick orange boxes.
First, the aforementioned relevance probabilities are estimated with a \emph{learned} gating GCN, allowing global structure to be taken into account. Second, the model discriminatively re-ranks the top-$K$ queries output by the generative decoder.

\noindent
\textbf{Global gating}
\newcite{bogin2019gnn} showed that an oracle relevance probability can increase model performance, but computed $\rho_v$ from local information only.

We propose to train a GCN to directly predict
$\rho_v$ from the global context of the question and DB.

The input to the gating GCN is the same graph described in \S\ref{sec:base_model}, except we add a new node $v_\text{global}$, connected to all other nodes with a special edge type.
To predict the question-conditioned relevance of a node, we need a representation for both the DB constant and the question. Thus, we define the input to the GCN at node $v$ to be $\bm{g}^{(0)}_v = FF([\bm{r}_v;\bar{\bm{h}}_v; \rho_v])$, where '$;$' is concatenation, $FF(\cdot)$ is a feed-forward network, and $\bm{\bar{h}_v} = \sum_i  p_{\text{link}}(x_i \mid v) \cdot \bm{e}_i$ is a weighted average of contextual representations of question tokens.
The initial embedding of $v_\text{global}$ is randomly initialized. A relevance probability is computed per DB constant based on the final graph representation: $\rho_v^{\text{global}} = \sigma(FF(\bm{g}^{(L)}_v))$. This probability replaces $\rho_v$ at the input to the encoder GCN (Figure~\ref{fig:highlevel}).

Because we have the gold query $y$ for each question, we can extract the gold subset of DB constants $\sU_y$, i.e., all DB constants that appear in $y$. We can now add a \emph{relevance loss} term $-\sum_{v \in \sU_y} \log \rho_v^\text{global} - \sum_{v \notin \sU_y} \log (1 - \rho_v^\text{global})$ to the objective. Thus, the parameters of the gating GCN are trained from the relevance loss and the usual \emph{decoding loss}, a ML objective over the gold sequence of decisions that output the query $y$.

\noindent
\textbf{Discriminative re-ranking}
Global gating provides a more accurate model for softly predicting the correct subset of DB constants. However, parsing is still auto-regressive and performed with a local similarity function. To overcome this, we separately train a discriminative model \cite{collins2005discriminative,Raymond:2006:DRS:1273073.1273107,lu-etal-2008-generative,fried-etal-2017-improving} to re-rank the top-$K$ queries in the decoder's output beam. The re-ranker scores each candidate tuple $(x, S, \hat{y})$, and thus can globally reason over the entire candidate query $\hat{y}$.

We focus the re-ranker capacity on the main pain point of zero-shot parsing -- the set of DB constants $\sU_{\hat{y}}$ that appear in $\hat{y}$.
At a high-level (Figure~\ref{fig:reranker}), for each candidate we compute a logit 
$s_{\hat{y}} = \bm{w}^\top FF(\bm{f}_{\sU_{\hat{y}}}, \bm{e}^\text{align})$, where $\bm{w}$ is a learned parameter vector, $\bm{f}_{\sU_{\hat{y}}}$ is a representation for the set $\sU_{\hat{y}}$, and $\bm{e}^\text{align}$ is a representation for the global alignment between question words and DB constants. The re-ranker is trained to minimize the \emph{re-ranker loss}, the negative log probability of the correct query $y$. We now describe the computation of $\bm{f}_{\sU_{\hat{y}}}$ and $\bm{e}^\text{align}$, based on a re-ranking GCN.

Unlike the gating GCN, the re-ranking GCN takes as input only the sub-graph induced by the selected DB constants $\sU_{\hat{y}}$, and the global node $v_\text{global}$. The input is represented by $\bm{f}^{(0)}_v = FF(\bm{r}_v; \bar{\bm{h}}_v)$, and after L propagation steps we obtain $\bm{f}_{\sU_{\hat{y}}} = \bm{f}^{(L)_{v_\text{global}}}$. Note that the global node representation is used to describe and score the question-conditioned sub-graph, unlike the gating GCN where the global node mostly created shorter paths between other graph nodes.

The representation $\bm{f}_{\sU_{\hat{y}}}$ captures global properties of selected nodes but ignores nodes that were not selected and are possibly relevant. Thus, we compute a representation $\bm{e}^\text{align}$, which captures whether question words are aligned to selected DB constants. We define a representation for every node $v \in \sV$:
\[
\bm{\varphi}_v = 
\begin{cases}
  \bm{f}_v^{(L)} & \text{ if } v \in \sU_{\hat{y}} \\
  \bm{r}_v & \text{otherwise}
\end{cases}
\]
Now, we compute for every question word $x_i$ a representation of the DB constants it aligns to: $\bm{l}^\varphi_i = \sum_{v \in \sV} p_\text{link}(v \mid x_i) \cdot \bm{\varphi}_v $. We concatenate this representation to every word $\bm{e}_i^{\text{align}} = [\bm{e}_i; \bm{l}^\varphi_i]$, and compute the vector $\bm{e}^\text{align}$ using attention over the question words, where the attention score for every word is $\bm{e}_i^{\text{align}\top} \bm{w}_\text{att}$ for a learned vector $\bm{w}_\text{att}$. The goal of this term is to allow the model to recognize whether there are any attended words that are aligned with DB constants, but these DB constants were not selected in $\sU_{\hat{y}}$.

In sum, our model adds a gating GCN trained to softly select relevant nodes for the encoder, and a re-ranking GCN that globally reasons over the subset of selected DB constants, and captures 
whether the query properly covers question words.
\section{Experiments and Results}

\begin{table}[t]
\centering
{\small
\begin{tabular}{|l l|}
\hline
Model                     & Accuracy \\ \hline
\textsc{SyntaxSQLNet}               & 19.7\% \\
\textsc{GNN}                    & 39.4\%  \\ \hline
\textsc{\textbf{Global-GNN}}         & \textbf{47.4\%} \\
\hline
\end{tabular}
}
\caption{Test set accuracy of \textsc{Global-GNN} compared to prior work on \spider{}.}
\label{tab:results_test}
\end{table}

\begin{table}[t]
\centering
{\scriptsize
\begin{tabular}{|l l l l l|}
\hline
Model                     & Acc. & Beam & \textsc{Single} & \textsc{Multi} \\ \hline
\textsc{SyntaxSQLNet}               & 18.9\% & & 23.1\% & 7.0\% \\ 
\textsc{GNN}                    & 40.7\%  & & 52.2\% & 26.8\% \\
+ \textsc{Re-implementation}                    & 44.1\% & 62.2\% & 58.3\% & 27.6\% \\ \hline
\textsc{\textbf{Global-GNN}}         & \textbf{52.1\%} & \textbf{65.9 \%} & \textbf{61.6}\% & \textbf{40.3}\% \\ 
- \textsc{No Global Gating}                       & 48.8\% & 62.2\% & 60.9\% & 33.8\% \\ 
- \textsc{No Re-ranking}                       & 48.3\%  & \textbf{65.9\%} & 58.1\% & 36.8\% \\ 
- \textsc{No Relevance Loss}                       & 50.1\%  & 64.8\%  & 60.9\% & 36.6\% \\ 
\textsc{No Align Rep.}                       & 50.8\% & \textbf{65.9\%} & 60.7\% & 38.3\% \\
\textsc{Query Re-ranker}                       & 47.8\%  & \textbf{65.9\%} & 55.3\% & 38.3\% \\ 
\hline
\textsc{Oracle Relevance}  & 56.4\% & 73.5\% & & \\ 
\hline
\end{tabular}
}
\caption{Development set accuracy for various experiments. The column `Beam' indicates the fraction of examples where the gold query is in the beam ($K=10$).}
\label{tab:results}
\end{table}

\noindent
\textbf{Experimental setup}
We train and evaluate on 
\spider{} \cite{yu2018spider}, which contains 7,000/1,034/2,147 train/development/test examples, using the same pre-processing as \citet{bogin2019gnn}. To train the re-ranker, we take $K=40$ candidates from the beam output of the decoder. At each training step, if the gold query is in the beam, we calculate the loss on the gold query and 10 randomly selected negative candidates. At test time, we re-rank the best $K=10$ candidates in the beam, and break re-ranking ties using the auto-regressive decoder scores (ties happen since the re-ranker considers the DB constants only and not the entire query).
We use the official \spider{} script for evaluation, which tests for loose exact match of queries.

\noindent
\textbf{Results}
As shown in Table \ref{tab:results_test}, the accuracy of our proposed model (\textsc{Global-GNN}) on the hidden test set is 47.4\%, 8\% higher than current state-of-the art of 39.4\%. Table \ref{tab:results} shows accuracy results on the development set for different experiments. We perform minor modifications to the implementation of \newcite{bogin2019gnn}, improving the accuracy from 40.7\% to 44.1\% (details in appendix \ref{app:reimpl}). We follow \newcite{bogin2019gnn}, measuring accuracy on easier examples where queries use a single table (\textsc{Single}) and those using more than one table (\textsc{Multi}).

\textsc{Global-GNN} obtains 52.1\% accuracy on the development set, substantially higher than all previous scores. Importantly, the performance increase comes mostly from queries that require more than one table, which are usually more complex.

Removing any of our two main contributions (\textsc{No Global Gating}, \textsc{No Re-ranking}) leads to a 4\% drop in performance. Training without the relevance loss (\textsc{No Relevance Loss}) results in a 2\% accuracy degrade. Omitting the representation $e^{\text{align}}$ from the re-ranker (\textsc{No Align Rep.}) reduces performance, showing the importance of identifying unaligned question words. 

We also consider a model that ranks the entire query and not only the set of DB constants.
We re-define $s_{\hat{y}} = \bm{w}^\top FF(\bm{f}_{\sU_{\hat{y}}}, \bm{h}^\text{align}, \bm{h}^\text{query})$, where $\bm{h}^\text{query}$
is a concatenation of the last and first hidden states of a BiLSTM run over the output SQL query (\textsc{Query Re-ranker}).
We see performance is lower, and most introduced errors are minor mistakes such as \texttt{min} instead of \texttt{max}. This shows that our re-ranker excels at choosing DB constants, while the decoder is better at determining the SQL query structure and the SQL logical constants.

Finally, we compute two oracle scores to estimate future headroom. Assuming a perfect global gating, which gives probability $1.0$ iff the DB constant is in the gold query, increases accuracy to 63.2\%. Adding to that a perfect re-ranker leads to an accuracy of 73.5\%.

\noindent
\textbf{Qualitative analysis}
Analyzing the development set, we find two main re-occurring patterns, where the baseline model is wrong, but our parser is correct. (a) \emph{coverage}: when relevant question words are not covered by the query, which results in a missing joining of tables or selection of columns (b)  \emph{precision}: when unrelated tables are joined to the query due to high lexical similarity. Selected examples are in Appendix \ref{app:examples}.

\noindent
\textbf{Error analysis} In 44.4\% of errors where the correct query was in the beam, the selection of $\sU$ was correct but the query was wrong. Most of these errors are caused by minor local errors, e.g., \texttt{min}/\texttt{max} errors, while the rest are due to larger structural mistakes, indicating that a global model that jointly selects both DB constants and SQL tokens might further improve performance. Other types of errors include missing or extra columns and tables, especially in complex queries.
\section{Conclusion}

In this paper, we demonstrate the importance of global decision-making for zero-shot semantic parsing, where 
selecting the relevant set of DB constants is challenging.
We present two main technical contributions. First, we use a gating GCN that globally attends the input question and the entire DB schema to softly-select the relevant DB constants. Second, we re-rank the output of a generative semantic parser by globally scoring the set of selected DB-constants. Importantly, these contributions can be applied to any zero-shot semantic parser with minimal modifications. Empirically, we observe a substantial improvement over the state-of-the-art on the \spider{} dataset, showing the effectiveness of both contributions.

\section*{Acknowledgments}
This research was partially supported by The
Yandex Initiative for Machine Learning. This work was completed in partial fulfillment for the Ph.D degree of the first author.

\bibliography{all}
\bibliographystyle{acl_natbib}

\appendix

\clearpage

\section{Re-implementation}
\label{app:reimpl}

We perform a simple modification to \newcite{bogin2019gnn}. We add \emph{cell values} to the graph, in a similar fashion to \newcite{krishnamurthy2017neural}. Specifically, we extract the cells of the first 5000 rows of all tables in the schema, during the pre-processing phase. We then consider every cell $q$ of a column $c$, which has a partial match with any of the question words $(x_1, \dots, x_{|x|})$. We then add nodes representing these cells to all of the model's graphs, with extra edges $(c, q)$ and $(q, c)$.

\section{Selected examples}
\label{app:examples}
Selected examples are given in Table~\ref{tab:examples}.
\begin{table*}[!t]
\centering
\footnotesize
\begin{tabular}{L{1.3cm}p{2.8cm}L{3cm}L{7.1cm}}
\toprule
{\bf Category} & {\bf Question} & {\bf Schema} & {\bf Predicted Queries}\\
 \midrule
  Coverage & Show the name of the teacher for the math course. & 
  \textbf{course}: course\_id, staring\_date, course  \newline
\textbf{teacher}: teacher\_id, name, age, hometown \newline
\textbf{course\_arrange}: course\_id, teacher\_id, grade    & 
  \textbf{Baseline: } \texttt{SELECT teacher.name FROM teacher WHERE teacher.name = 'math'} \newline 
  \textbf{Our Model: } \texttt{SELECT teacher.name FROM teacher JOIN course\_arrange ON teacher.teacher\_id = course\_arrange.teacher\_id JOIN course ON course\_arrange.course\_id = course.course\_id WHERE course.course = 'math'} 
 \\\midrule
  Coverage & Who is the first student to register? List the first name, middle name and last name. & 
  \textbf{students}: student\_id, current\_address\_id, first\_name, middle\_name, last\_name, ... \newline
\textbf{student\_enrolment}: student\_enrolment\_id, degree\_program\_id, ...  \newline
...& 
  \textbf{Baseline: } \texttt{SELECT students.first\_name, students.middle\_name, students.last\_name FROM students} \newline 
  \textbf{Our Model: } \texttt{SELECT students.first\_name, students.middle\_name, students.last\_name FROM students ORDER BY students.date\_first\_registered LIMIT 1} 
 \\\midrule
  Coverage &  List all singer names in concerts in year 2014. & 
\textbf{singer}: singer\_id, name, country, song\_name, song\_release\_year, ... \newline
\textbf{concert}: concert\_id, concert\_name, year, ... \newline
\textbf{singer\_in\_concert}: concert\_id, singer\_id  \newline  ... & 
  \textbf{Baseline: } \texttt{SELECT singer.name FROM singer WHERE singer.song\_release\_year = 2014} \newline 
  \textbf{Our Model: } \texttt{SELECT singer.name FROM singer JOIN singer\_in\_concert ON singer.singer\_id = singer\_in\_concert.singer\_id JOIN concert ON singer\_in\_concert.concert\_id = concert.concert\_id WHERE concert.year = 2014} 
 \\\midrule
  Precision & What are the makers and models? & 
  \textbf{car\_makers}: id, maker, fullname, country \newline
\textbf{model\_list}: modelid, maker, model \newline ...   & 
  \textbf{Baseline: } \texttt{SELECT car\_makers.maker, model\_list.model FROM car\_makers JOIN model\_list ON car\_makers.id = model\_list.maker} \newline 
  \textbf{Our Model: } \texttt{SELECT model\_list.maker, model\_list.model FROM model\_list} 
 \\\midrule
  Precision & Return the id of the document with the fewest paragraphs. &
\textbf{documents}: document\_id, document\_name, document\_description, ... \newline
\textbf{paragraphs}: paragraph\_id, document\_id, paragraph\_text, ... \newline
...& 
  \textbf{Baseline: } \texttt{SELECT documents.document\_id FROM documents JOIN paragraphs ON documents.document\_id = paragraphs.document\_id GROUP BY documents.document\_id ORDER BY COUNT(*) LIMIT 1} \newline 
  \textbf{Our Model: } \texttt{SELECT paragraphs.document\_id FROM paragraphs GROUP BY paragraphs.document\_id ORDER BY COUNT(*) ASC LIMIT 1} 
  \\
 \bottomrule
\end{tabular}
\caption{Selected correct examples where the baseline model is wrong, but our parser is correct.}
\label{tab:examples}
\end{table*}

\end{document}